\documentclass{article}

\usepackage[verbose=true,letterpaper]{geometry}
\AtBeginDocument{
  \newgeometry{
    textheight=9in,
    textwidth=6in,
    top=1in,
    headheight=12pt,
    headsep=25pt,
    footskip=30pt
  }  
}

\usepackage[utf8]{inputenc} 
\usepackage[T1]{fontenc}    
\usepackage{hyperref}       
\usepackage{url}            
\usepackage{booktabs}       
\usepackage{amsfonts}       
\usepackage{nicefrac}       
\usepackage{microtype}      
\usepackage{xcolor}         
\usepackage{graphicx}
\usepackage{multirow}
\usepackage{amsmath, calc}
\usepackage{dirtytalk}
\usepackage{algorithm2e}
\usepackage{algpseudocode}
\usepackage{authblk}
\usepackage{xr}

\title{Predicting the Next Action by Modeling the Abstract Goal}

\author[1]{Debaditya Roy}
\author[1,2]{Basura Fernando}
\affil[1]{Institute of High Performance Computing (IHPC), Agency for Science, Technology and Research (A*STAR), 1 Fusionopolis Way, \#16-16 Connexis, Singapore 138632, Republic of Singapore.}
\affil[2]{Centre for Frontier AI Research (CFAR), Agency for Science, Technology and Research (A*STAR), 1 Fusionopolis Way, \#16-16 Connexis, Singapore 138632, Republic of Singapore.}
\date{} 

\begin{document}
\maketitle

\begin{abstract}
The problem of anticipating human actions is an inherently uncertain one. 
However, we can reduce this uncertainty if we have a sense of the goal that the actor is trying to achieve.
Here, we present an action anticipation model that leverages goal information for the purpose of reducing the uncertainty in future predictions.
Since we do not possess goal information or the observed actions during inference, we resort to visual representation to encapsulate information about both actions and goals. Through this, we derive a novel concept called abstract goal which is conditioned on observed sequences of visual features for action anticipation.
{We design the abstract goal as a distribution whose parameters are estimated using a variational recurrent network. 
We sample multiple candidates for the next action and introduce a goal consistency measure to determine the best candidate that follows from the abstract goal.}
Our method obtains impressive results on the very challenging Epic-Kitchens55 (EK55), EK100, and EGTEA Gaze+ datasets.
We obtain absolute improvements of \textbf{+13.69},  \textbf{+11.24}, and \textbf{+5.19} for Top-1 verb, Top-1 noun, and Top-1 action anticipation accuracy respectively over prior state-of-the-art methods for seen kitchens (S1) of EK55.
Similarly, we also obtain significant improvements in the unseen kitchens (S2) set for Top-1 verb (\textbf{+10.75}), noun (\textbf{+5.84}) and action (\textbf{+2.87}) anticipation. 
Similar trend is observed for EGTEA Gaze+ dataset, where absolute improvement of \textbf{+9.9}, \textbf{+13.1} and \textbf{+6.8} is obtained for noun, verb, and action anticipation.
It is through the submission of this paper that our method is currently the new state-of-the-art for action anticipation in EK55 and EGTEA Gaze+. 
\url{https://competitions.codalab.org/competitions/20071#results}.

Code available at \url{https://github.com/debadityaroy/Abstract_Goal}
\end{abstract}

\section{Introduction}
Human action anticipation is an important problem having applications in human-robot collaboration, smart houses, assistive robotics and wearable virtual assistants.
Action anticipation models aim to predict the most plausible future action that is going to happen in the immediate future.
Humans are rational and actions taken by us take us closer to \emph{what we would like to attain at the end}, i.e. our goal.
After we identify the goal, humans formulate a plan to execute actions to achieve that goal as explained in the \emph{seven stages of action cycle}~\cite{shove2007design}.
Therefore, to predict what a person is going to do next accurately, it is useful to infer their goals. 
Besides, it has been shown that it is possible to infer the goal of the person from observing their actions~\cite{baker2007goal}.
However, action anticipation literature~\cite{furnari2020rolling,girdhar2021anticipative,qi2021self,roy2021action,sener2020temporal,wang2021interactive,wu2021Anticipate,zatsarynna2021multi} has not made use of this vital information that governs human actions or it seems that goal modeling is not a popular solution.

Goal inference from observed actions (or features) is an extremely challenging task. 
Two different goals can share the same partial action sequences while different persons may have different execution plans for the same goal. 
Therefore, goal modeling is a highly stochastic problem. 
In this paper, we make use of a stochastic method~\cite{chung2015recurrent,fraccaro2016sequential} for goal modeling to improve action anticipation {that goes beyond the deterministic latent goal representation introduced in~\cite{Roy_2022_WACV}}.
Interestingly, if we know the goal of the person, then it reduces the uncertainty when predicting future actions that they might take to accomplish the goal as backed up by the cognitive science literature~\cite{baker2009action,baker2007goal,shove2007design}.
{Our approach is also motivated by procedure planning~\cite{chang2020procedure} where a sequence of intermediate actions is predicted based on goals provided as the final visual representation or the overall activity~\cite{miech2019howto100m}.}

\begin{figure}[t]
\centering    
\includegraphics[width=\linewidth, height=130pt]{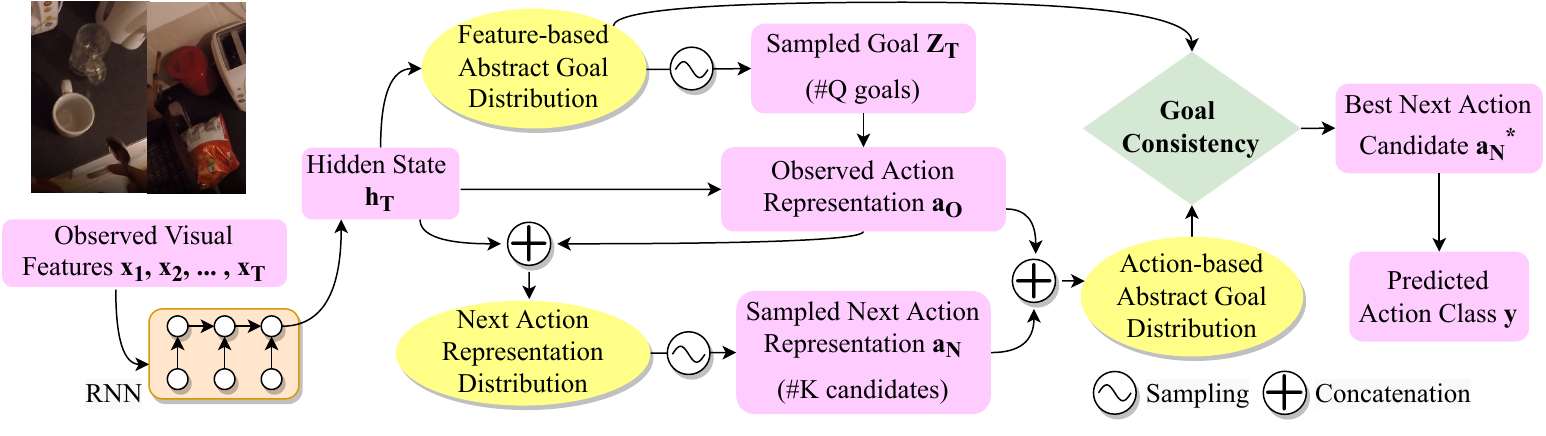}
    \caption{Illustration of the model design for abstract goal-based action anticipation. Yellow ellipses represent distributions and pink boxes represent various variables of the model. Best viewed on a screen.}
    \label{motivation}
\end{figure}

Explicit goal inference is not trivial in action anticipation as the goal of the activity is not available as ground truth.
To this end, we outline our approach in Figure~\ref{motivation}.
We learn a latent distribution from the observed visual features using stochastic RNN \cite{chung2015recurrent} which we call ``feature-based abstract goal'' distribution.
Given the hidden state of RNN and a sampled feature-based abstract goal ($\mathbf{z_T}$ in Figure~\ref{motivation}), we obtain a representation for the observed action ($\mathbf{a_O}$). 
Afterward, we model the next action-representation distribution that is conditioned on the sampled feature-based abstract goal and the observed action-representation and sample a candidate's next action-representation ($\mathbf{a_N}$) from it. 
Given the observed and next action representations, we learn the distribution of the ``action-based abstract goal'' using the generative variational framework. 
Therefore, we infer two kinds of \emph{abstract goals} - one using visual feature sequence and another using \emph{action-representations} unlike ~\cite{abu2019uncertainty,mehrasa2019variational} that use a single latent distribution to model past actions.

The setup of next action anticipation assumes that the next action in the sequence can be reliably inferred from the observed action(s). 
Hence, the feature-based abstract goal distribution derived from observed features and action-based abstract goal distribution derived from next action representation should be consistent.
The action that is most likely to happen in the future (“next best action”) is the one that maximizes consistency between the two abstract goal distributions.
We sample multiple next action-representation candidates and introduce a new goal consistency criterion to measure the suitability of each candidate when predicting the next action--see Figure~\ref{motivation}.
During learning, we also use goal consistency as a loss function to obtain a better model.
Such a mechanism is not present in previous stochastic approaches ~\cite{abu2019uncertainty,mehrasa2019variational} which only minimizes KL divergence between prior and posterior latent distribution to obtain the best samples.
We show that goal consistency has the biggest impact on action anticipation. 
  
Our approach yields large improvements when predicting the next action in unscripted activities from the Epic-Kitchens55 and EGTEA Gaze+ datasets.
We also obtain significantly better results for unseen kitchens of the Epic-Kitchens100. 
Our contributions are:
(1) a novel stochastic abstract goal concept for action anticipation. (2) a novel stochastic model to learn abstract goal distributions and then use it for effective action anticipation in unscripted activities. (3) a novel goal consistency term that measures how well a plausible future action (next action) aligns with abstract goal distributions. 
The code will be made publicly available.

\section{Related work}
\label{sec.rel}
Research in action anticipation is gaining popularity in recent years thanks to progress in datasets~\cite{damen2021rescaling} and challenges~\cite{epic}. 

\subsection{Goals in action anticipation}
The activity label of the entire action sequence is used to anticipate the next action in~\cite{sener2020temporal}.
In ~\cite{Roy_2022_WACV}, observed features are used to obtain a fixed latent goal from visual features.
However, humans often pursue multiple goals simultaneously~\cite{damen2021rescaling}. 
Hence, we propose an abstract goal distribution that serves as a representation for one or more underlying intention(s).
The final visual representation is considered as the goal in~\cite{chang2020procedure}.
Our abstract goal allows us to anticipate actions without any knowledge of the overall activity label~\cite{sener2020temporal} or final visual representation~\cite{chang2020procedure}.

\subsection{Features for action anticipation}
Multiple representations of such as spatial, motion, and bag of objects are used to predict future actions in~\cite{furnari2020rolling}. 
The authors showed that unrolling an LSTM for multiple time steps in the future is beneficial for the next action prediction. 
In~\cite{roy2021action}, human-object interactions are encoded as features and fed to a transformer encoder-decoder to predict the features of future frames and the corresponding future actions. 
In~\cite{liu2020forecasting}, spatial attention maps of future human-object interactions are estimated to predict the next action.
%

In~\cite{unlabeled_anticipation}, multiple future visual representations are generated from an input frame using a CNN and the future action is predicted. 
Similarly in~\cite{wu2021Anticipate}, an RNN is used to generate the intermediate frames between the observed frames and the anticipated action. 
In \cite{ke2019time}, temporal features are computed using time-conditioned skip connections to anticipate the next action. 
In~\cite{when, Loh_2022_CVPR}, RNNs are used to predict future actions conditioned on observed action labels.
Similarly, in~\cite{gong2022future}, a transformer is used to encode past actions and duration while another transformer decoder is used to predict both future actions and their duration.
In~\cite{girdhar2021anticipative}, every frame is divided into patches and combined using the Visual Transformer (ViT)~\cite{dosovitskiy2020image} to obtain a frame representation. 
These frame representations are combined using a temporal transformer to predict future features and action labels.
In~\cite{wu2022memvit}, long-range sequences are summarized by processing smaller temporal sequences using multi-scale ViTs and caching them in memory as context. 
Each context is attended hierarchically in time using multiple temporal transformers for action anticipation. 
{We use a Gated Recurrent Unit (GRU) to summarize frame features and learn feature-based abstract goal distribution parameters to generate observed and next action representations.}

\subsection{Past-Future Correlation for anticipation}
An action anticipation model that correlates past observed features with the future using Jaccard vector similarity is presented in~\cite{fernando2021}.
In~\cite{nmns} proposes a neural memory network to compare an input (spatial representation or labels) with the existing memory content to predict future action labels.
Similarly, \cite{qi2021self} proposes an action anticipation framework with a self-regulated learning process. They correlate similarities and differences between past and current frames. 
In~\cite{leveraging}, a predictive model directly anticipates the future action and a low-rank linear transitional model predicts the current action and correlates with the predicted future actions. 
Similarly, counterfactual reasoning is used to improve action anticipation in~\cite{zhangif21}.
{Our approach correlates the past and future by enforcing goal consistency between the two abstract goal distributions computed using observed features and the next action.}
Although our method can be also used for long term action forecasting~\cite{Ng2020} and early action recognition~\cite{shi2018action,sadegh2017encouraging}, in this work we primarily focus on action anticipation.

\section{Action anticipation with abstract goals}


In this section we explain our model design outlined in Figure~\ref{motivation} and then describe the related work in Section~\ref{sec.rel}.

\subsection{Feature-based abstract goal}
\label{sec:abs_goal}
In this section we describe how to generate \emph{feature-based abstract goal} representation using variational RNN (VRNN) framework~\cite{chung2015recurrent,fraccaro2016sequential}.
Let us denote the observed feature sequence by $\mathbf{x}_1, \mathbf{x_2}, \cdots, \mathbf{x}_T$ where $\mathbf{x}_t \in \mathbb{R}^{d_f}$.
Following standard VRNN, a Gaussian distribution  $q_t=\mathcal{N}(\boldsymbol{\boldsymbol{\mu}}_{t,prior},\boldsymbol{\boldsymbol{\sigma}}_{t,prior})$ is used to model the prior distribution of the abstract goal ($\mathbf{z_t}$) given the observed feature sequence $t \in \{1, \cdots, T\}$ where $\boldsymbol{\mu}_{t,prior},\boldsymbol{\boldsymbol{\sigma}}_{t,prior} \in \mathbb{R}^{d_z}$.
The parameters of abstract goal distribution $q_t$ are estimated using an MLP denoted by $\phi_{prior} : \mathbb{R}^{d_h}\to\mathbb{R}^{d_z}$) using the hidden state of a RNN ($\mathbf{h}_{t-1} \in \mathbb{R}^{d_h}$) learned from the previous $t-1$ features as follows: 
\begin{align}
q(\mathbf{z}_{t} | \mathbf{x}_{1:t-1}) \sim \mathcal{N}(\boldsymbol{\boldsymbol{\mu}}_{t,prior}, \boldsymbol{\boldsymbol{\sigma}}_{t,prior}), \\ \text{ where } (\boldsymbol{\boldsymbol{\mu}}_{t,prior}, \boldsymbol{\boldsymbol{\sigma}}_{t,prior}) = \phi_{prior}(\mathbf{h}_{t-1}).
\label{prior}
\end{align}
Note that there are two separate MLPs, one to obtain $\boldsymbol{\boldsymbol{\mu}}_{t,prior}$ and another to obtain $\boldsymbol{\boldsymbol{\sigma}}_{t,prior}$.
We apply $softplus$ activation to the $\phi_{prior}$ network that estimates the standard deviation ($\boldsymbol{\boldsymbol{\sigma}}_{t,prior}$). Unless otherwise specified, all $\phi_{<>}(\cdot)$ used in our model are two layered neural networks with ReLU activation. 

The posterior distribution of the abstract goal ($r$) computes the effect of observing the incoming new feature $\mathbf{x}_{t}$ as follows:
\begin{align} 
r(\mathbf{z}_{t}|  \mathbf{x}_{1:t}) \sim \mathcal{N}(\boldsymbol{\boldsymbol{\mu}}_{t,pos}, \boldsymbol{\boldsymbol{\sigma}}_{t,pos}).
\label{encoder}
\end{align}
As before, the parameters of $r$ are computed by a two layered MLP $\phi_{pos}: \mathbb{R}^{2 \times d_z} \to \mathbb{R}^{d_z}$ using both the last hidden state of RNN ($\mathbf{h}_{t-1}$) and the incoming feature ($\mathbf{x}_t$) as follows:
\begin{align} 
\boldsymbol{\boldsymbol{\mu}}_{t,pos}, \boldsymbol{\boldsymbol{\sigma}}_{t,pos} = \phi_{pos}([\phi_x(\mathbf{x}_t), \phi_h(\mathbf{h}_{t-1})]),
\label{encoder2}
\end{align}
where $\phi_x: \mathbb{R}^{d_f} \to \mathbb{R}^{d_z}$, $\phi_h: \mathbb{R}^{d_h} \to \mathbb{R}^{d_z}$ are linear layers and $[\cdot,\cdot]$ represents vector concatenation.
We use the reparameterization trick~\cite{kingma2013auto} to sample an abstract goal ($\mathbf{z}_{t} \in\mathbb{R}^{d_z}$) from the prior distribution $q(\mathbf{z}_{t}| \mathbf{x}_{1:t-1})$ as follows:
\begin{equation}
\mathbf{z}_{t} = \boldsymbol{\boldsymbol{\mu}}_{t,prior} + \boldsymbol{\boldsymbol{\sigma}}_{t,prior} \odot \boldsymbol{\epsilon},
\label{reparam}
\end{equation}
where $\boldsymbol{\epsilon} \sim \mathcal{N}(\mathbf{0},\mathbf{1})\in \mathbb{R}^{d_z}$ is a standard Gaussian distribution. 
Then sampled $\mathbf{z}_{t}$ is used to obtain the next hidden state of the RNN\footnote{Our RNN is a standard GRU cell.} as follows:
\begin{equation}
\mathbf{h}_t = RNN(\mathbf{h}_{t-1}, [\phi_x(\mathbf{x}_t), \phi_z(\mathbf{z}_{t})]), \forall t \in {1,  \cdots,
T}
\label{rnn}
\end{equation}
where $\phi_z: \mathbb{R}^{d_z} \to \mathbb{R}^{d_z}$ acts as a feature extractor over $\mathbf{z}_{t}$. 
The sampled abstract goal ($\mathbf{z}_{t}$) can be used to reconstruct (or generate) the feature sequence as done in VRNN framework~\cite{chung2015recurrent,fraccaro2016sequential}. However, we use it to represent feature-based abstract goal. Our intuition comes from the fact that humans derive action plans from goals, and videos are a realization of this action plan. Therefore, by construction, goal determines the video (feature evolution in our case). Interestingly, as the abstract goal latent variable encapsulates the video feature generation process, by analogical similarity, we make the proposition that latent variable ($\mathbf{z}_{t}$) represents the notion of feature-based abstract goal.

Therefore, we denote the \say{feature-based abstract goal distribution}  as follows:
\begin{equation}
p(\mathbf{z}_T) = q(\mathbf{z}_{T}| \mathbf{x}_{1:T-1}).
\label{eq.feat.goal}
\end{equation} 
The abstract goal distribution represents all abstract goals with respect to a particular observed features sequence. Any observed action may lead to more than one goal. Our abstract goal representation captures these variations.

%
%
\subsection{Action representations}
Human actions are causal in nature and the next action in a sequence depends on the earlier actions.
For example, \textit{washing vegetables} is succeeded by \textit{cutting vegetables} when the goal is \say{making a salad}.
We capture the causality between observed and next actions using  ``the observed action representation'' and the ``next action representation''.
We obtain the \textbf{observed action representation} ($\mathbf{a}_O$) using feature-based abstract goal and the hidden state of RNN as follows:
\begin{equation}
    \mathbf{a}_O = \phi_O([\phi_z(\mathbf{z}_T), \phi_h(\mathbf{h}_T)]).
    \label{eq.ao}
\end{equation}
Here $\phi_O: \mathbb{R}^{2 \times d_z} \to \mathbb{R}^{d_h}$ and $\mathbf{z}_T$ is sampled from the abstract goal distribution $p(\mathbf{z}_T)$ using Equation~\ref{reparam}.

Then we obtain the distribution of \textbf{next action representation} ($\mathbf{a}_N$) conditioned on the hidden state of the RNN and the observed action representation denoted by $p(\mathbf{a}_N | \mathbf{h}_T, \mathbf{a}_O)$.
The reason for modeling next action representation as a distribution conditioned on hidden state and the observed action representation is twofold. First, a particular observed action may lead to different next actions depending on the context and goal.
Note that in our model, both observed action representation $\mathbf{a}_O$ and the RNN hidden state $\mathbf{h}_T$ depend on the feature-based abstract goal representation.
Second, there can be variations in human behavior when executing the same task.
The next action representations are generated using a Gaussian distribution $\mathcal{N}(\boldsymbol{\mu}_{\mathbf{a}_N}, \boldsymbol{\sigma}^2_{\mathbf{a}_N})$ where $\boldsymbol{\mu}_{\mathbf{a}_N}, \boldsymbol{\sigma}^2_{\mathbf{a}_N} \in \mathbb{R}^{d_z} $ whose parameters are estimated as follows:

\begin{align}
    p(\mathbf{a}_N | \mathbf{h}_T, \mathbf{a}_O) \sim \mathcal{N}(\boldsymbol{\mu}_{\mathbf{a}_N}, \boldsymbol{\sigma}^2_{\mathbf{a}_N}) \nonumber \\ \text{ where }
    (\boldsymbol{\mu}_{\mathbf{a}_N}, (\boldsymbol{\sigma}_{\mathbf{a}_N})) = \phi_N([\phi_h(\mathbf{h}_T), \phi_a(\mathbf{a}_O)]),
    \label{eq.next.an}
\end{align}
where $\phi_a: \mathbb{R}^{d_h} \to  \mathbb{R}^{d_z} $ and  $\phi_N: \mathbb{R}^{2 \times d_z} \to \mathbb{R}^{d_z}$ are two separate MLPs. 
Now we sample multiple next action representations from the next action representation distribution using the reparameterization trick as in Equation~\ref{reparam2},
\begin{equation}
\mathbf{a}_N = \boldsymbol{\mu}_{\mathbf{a}_N} + \boldsymbol{\sigma}_{\mathbf{a}_N} \odot \boldsymbol{\epsilon},
\label{reparam2}
\end{equation}
where $\boldsymbol{\epsilon} \sim \mathcal{N}(\mathbf{0},\mathbf{1})\in \mathbb{R}^{d_z}$ is a standard Gaussian distribution.

%
%
\subsection{Action-based abstract goal representation}
\label{sec:act_goal}
Now, we obtain \textit{action-based abstract goal} from observed and next action representations using generative variational framework~\cite{kingma2013auto}.
The distribution for action-based abstract goal is modeled with a Gaussian distribution conditioned on the next action representation denoted by $q(\mathbf{z}_{N}| \mathbf{a}_{N})$ whose parameters are computed as:
\begin{align} 
q(\mathbf{z}_{N}| \mathbf{a}_{N}) \sim \mathcal{N}(\boldsymbol{\mu}_{Nq}, \boldsymbol{\sigma}_{Nq}), \nonumber \\ \text{ where } (\boldsymbol{\mu}_{Nq}, \boldsymbol{\sigma}_{Nq}) = \phi_{Nq}(\phi_a(\mathbf{a}_{N})),
\label{prior2}
\end{align}
where $\boldsymbol{\mu}_{Nq}, \boldsymbol{\sigma}_{Nq} \in \mathbb{R}^{d_z}$ and  $\phi_{Nq}: \mathbb{R}^{d_h} \to \mathbb{R}^{d_z} $ is implemented with two MLPs.
On the other hand, parameters of the action-based abstract goal distribution conditioned on both observed and next action representation are given as follows:
\begin{align}
r(\mathbf{z}_{N} | \mathbf{a}_N, \mathbf{a}_O )\sim \mathcal{N}(\boldsymbol{\mu}_{Nr}, \boldsymbol{\sigma}_{Nr})
\nonumber \\ \text{ where }
(\boldsymbol{\mu}_{Nr}, \boldsymbol{\sigma}_{Nr}) = \phi_{Nr}([\phi_a(\mathbf{a}_N), \phi_a(\mathbf{a}_O)]) 
\label{posterior2}
\end{align}
where $\boldsymbol{\mu}_{Nr}, \boldsymbol{\sigma}_{Nr} \in \mathbb{R}^{d_z}$ and $\phi_{Nr}:\mathbb{R}^{d_h} \to \mathbb{R}^{d_z}$ is a dual headed MLP.
%
Finally, the \textbf{action-based abstract goal} distribution for the next action $p(\mathbf{z}_N)$ is given by the distribution
\begin{equation}
    p(\mathbf{z}_N) = q(\mathbf{z}_{N}|\mathbf{a}_N).
    \label{eq.action.goal}
\end{equation}
We use both feature-based and action-based abstract goal representation to find the \emph{best candidate for next action} as explained in next section.
It should be noted that while the $q(\mathbf{z}_{N}| \mathbf{a}_{N})$ only depends on $\mathbf{a}_{N}$, the  $r(\mathbf{z}_{N} | \mathbf{a}_N, \mathbf{a}_O )$ depends on both $\mathbf{a}_N$ and  $\mathbf{a}_O$. 

\subsection{Next action anticipation with goal consistency}
\label{sec:consistency}

Given a sampled feature-based abstract goal $\mathbf{z}_T$, we select the best next action representation $\mathbf{a}_{N}^{*}$ using the divergence between $p(\mathbf{z}_T)$ distribution (eq.~\ref{eq.feat.goal}) and $p(\mathbf{z}_N)$ distribution~(eq.~\ref{eq.action.goal}).
We call this divergence as the \textbf{goal consistency criterion}.
Given $\mathbf{z}_T$, observed action $\mathbf{a}_O$ and the next sampled action $\mathbf{a}_N$, the goal consistency criterion is derived from the symmetric KL-divergence between $p(\mathbf{z}_T)$ and $p(\mathbf{z}_N)$ as follows:
\begin{equation}
D(\mathbf{a}_N) =\frac{D_{KL}(p(\mathbf{z}_T) || p(\mathbf{z}_N))  + D_{KL}(p(\mathbf{z}_N) || p(\mathbf{z}_T))}{2}.
\label{symkld}
\end{equation}
We choose the best next action candidate (i.e. the anticipated action candidate representation) $\mathbf{a}_{N}^{*}$ that minimizes the goal consistency criterion. 
The rationale is that the best anticipated action should have an action-based abstract goal representation $p(\mathbf{z}_N)$ that aligns with the feature-based abstract goal distribution $p(\mathbf{z}_T)$. 
We use the following algorithm to find the best next action candidate $\mathbf{a}_{N}^{*}$.
%
%
\begin{algorithm}
\begin{algorithmic}[1]
\small
\caption{Best next action selection\label{alg}}
\State Sample feat-based abstract goal $\mathbf{z}_T$ from eq.~\ref{eq.feat.goal} $\rightarrow$ $\mathbf{z}_T \sim q(\mathbf{z}_{T}| \mathbf{x}_{1:T-1})$
\State Get observed action representation $\mathbf{a}_O$ (eq.~\ref{eq.ao})
\State Get next action representation distribution $p(\mathbf{a}_N | \mathbf{h}_T, \mathbf{a}_O)$ (eq.~\ref{eq.next.an})
\State Sample $K$ next action representations  $\mathcal{N} = \{\mathbf{a}_N^1, \cdots \mathbf{a}_N^K\} \sim p(\mathbf{a}_N | \mathbf{h}_T, \mathbf{a}_O)$
\State Best next action $\mathbf{a}_{N}^{*} = \text{argmin}_{\mathbf{a}_N^k \in  \mathcal{N}} D(\mathbf{a}_N^k)$
\end{algorithmic}
\end{algorithm}

Finally, we predict the anticipated action from the selected next action representation as follows:
\begin{equation}
    \hat{\mathbf{y}} = \phi_{c} (\mathbf{a}_{N}^{*})    
\end{equation}
where $\phi_{c}: \mathbb{R}^{d_z} \to \mathbb{R}^{d_c}$ is the MLP classifier and $\hat{\mathbf{y}}$ is the class score vector.
It should be noted that in Algorithm~\ref{alg}, we sample only one feature-based abstraction goal in line 1 of the algorithm. However, during training we sample $Q$ number of feature-based abstraction goals and for each of them we sample $K$ number of next action representations. In this case, we select the best candidate from all $K \times Q$ next action representation candidates using Equation~\ref{symkld}. So, the next best action is consistent and does not rely too much on sampling as long as we sample sufficient candidate next actions.

Even if the feature-based abstract goal $P(\mathbf{z}_T)$ is obtained from VRNN framework~\cite{chung2015recurrent,fraccaro2016sequential}, the formulation of action representations $\mathbf{a_O}$ and $\mathbf{a_N}$, action-based abstract goal $P(\mathbf{z}_N)$ and goal consistency criterion is drastically different from ~\cite{abu2019uncertainty,mehrasa2019variational}.
In~\cite{Roy_2022_WACV}, goal consistency is defined between latent goals before and after the action using a hard threshold. 
Instead, our goal consistency is a symmetric KL divergence between $p(\mathbf{z}_T)$ and $p(\mathbf{z}_N)$ distributions which aims to align the two abstract goal distributions.
This also results in a massive improvement in next action anticipation performance as shown in the experiments.

\subsection{Loss functions and training of our model}\label{training}
Our anticipation network is trained using a number of losses. 
In contrast to prior stochastic methods~\cite{mehrasa2019variational}~\cite{abu2019uncertainty}, we introduce three KL divergence losses, based on a) feature-based abstract goal, b). action-based abstract goal ($L_{NG}$), 3. goal-consistency loss ($L_{GC}$).
The first loss function is used to learn the parameters of the feature-based abstract goal distribution. 
We compute the KL-divergence between the conditional prior and posterior distributions for every feature in the observed feature sequence and minimize the sum given as follows:
\begin{equation}
\mathcal{L}_{OG} = \sum_{t=1}^T D_{KL} ( r(\mathbf{z}_{t} | \mathbf{x}_{1:t}) || q(\mathbf{z}_{t} | \mathbf{x}_{1:t-1}) ).
\label{kld_timestep}
\end{equation}
This is based on the intuition that the abstract goal should not change due to a new observed feature.
Our second loss arises when we learn the action-based abstract goal distribution. 
As in equation~\ref{kld_timestep} above, we compute the KL-divergence between $r(\mathbf{z}_{N} | \mathbf{a}_{N}^{*}, \mathbf{a}_O)$ and $q(\mathbf{z}_{N} | \mathbf{a}_{N}^{*})$ distributions of action-based abstract goal distributions as follows:
\begin{equation}
\mathcal{L}_{NG} = D_{KL}( r(\mathbf{z}_{N} | \mathbf{a}_{N}^{*}, \mathbf{a}_O)|| q(\mathbf{z}_{N} | \mathbf{a}_{N}^{*})) .
\label{kldtimestep}
\end{equation}
We denote the corresponding best action-based abstract goal distribution by $p(\mathbf{z}^*_{N}) = q(\mathbf{z}_{N}|\mathbf{a}_{N}^{*})$.
The intuition is same as before, the goal should not change because of the next best action $\mathbf{a}_{N}^{*}$. 
Furthermore, the feature-based and action-based abstract goal distributions should be aligned with respect to the selected next best action $\mathbf{a}_{N}^{*}$. 
Therefore, we minimize the symmetric KL-Divergence between the feature-based and best-action-based abstract goal distribution as follows:
\begin{align}
    \mathcal{L}_{GC} = &\frac{D_{KL}(p(\mathbf{z}_T) || p(\mathbf{z}^*_{N})+ D_{KL}(p(\mathbf{z}^*_{N}) || p(\mathbf{z}_T)}{2}.
\end{align}
We coin this loss as \textbf{goal consistency loss}.
This loss is based on $D({\mathbf{a}_N})$ in Equation~\ref{symkld} with the only difference being that $p(\mathbf{z}^*_{N}) = q(\mathbf{z}_{N}|\mathbf{a}_{N}^{*})$ is computed with respect to the selected best next action representation $\mathbf{a}_{N}^{*}$.
Finally, we have the cross-entropy loss for comparing the model's prediction $\mathbf{\hat{y}}$ with the ground truth one-hot label $\mathbf{y}$ as follows
\begin{align}
\mathcal{L}_{NA} = - \sum  \mathbf{y} \odot \log(\mathbf{\hat{y}}).
\label{cel}
\end{align}
The loss function to train the model is a combination of all losses given as follows:
\begin{equation}
    \mathcal{L}_{total} = \mathcal{L}_{OG} + \mathcal{L}_{NG} + \mathcal{L}_{GC} + \mathcal{L}_{NA}.
\end{equation}
We experimented with adding different weights to each loss but there is no significant difference in learning.
Therefore, we weight them equally.

\section{Experiments and results}

\subsection{Datasets, features, and training details}
We use three well known action anticipation datasets, \textit{EPIC-KITCHENS55}\cite{epic} (EK55), \textit{EPIC-KITCHENS100}\cite{damen2021rescaling} (EK100) and \textit{EGTEA Gaze+}\cite{egtea} to evaluate.
%
%
%
%
We validate our models using the TSN features obtained from RGB and optical flow videos, and bag of object features provided by~\cite{furnari2020rolling} for a fair comparison with existing approaches. 
Our base model has the following parameters: observed duration - 2 seconds, frame rate - 3 fps, RNN (GRU) hidden dimension $d_h=$ 256, abstract goal dimension $d_z=$ 128, number of sampled feature-based abstract goals ($Q=3$), number of next-action-representation candidates ($K=10$), and fixed anticipation time - 1s (following EK55 and EK100 evaluation server criteria), unless specified otherwise. 
We use a batch size of 128 videos and train for 15 epochs with a learning rate of 0.001 using Adam optimizer with weight decay (AdamW) in Pytorch. All our MLPs have 256 hidden dimensions.
\subsection{Comparison with state-of-the-art}
We compare the performance of Abstract Goal (our method) with current state-of-the-art approaches on both the seen and unseen test sets of EK55 datasets in Table~\ref{epic55} using a late fusion of TSN-RGB, TSN-Flow, and Object features like most of the prior work. 
We train separate models for verb and noun anticipation and combine their predictions to obtain action anticipation accuracy.
Our method outperforms all other prior state-of-the-art methods by a significant margin for both seen kitchens (S1) and unseen kitchens (S2). 
Notably, we outperform Transformer-based AVT~\cite{girdhar2021anticipative} and Temporal-Aggregation~\cite{sener2020temporal} in all measures in both seen and unseen kitchens except for Top-5 accuracy on unseen kitchens. 
A similar trend can be seen for EGTEA Gaze+ dataset in Table~\ref{egtea} where our method outperforms all compared methods. 
We believe this improvement is due to two factors, (i) stochastic modeling is massively important for action anticipation, and (ii) the effective use of goal information is paramount for better action anticipation. 

\begin{table*}
\resizebox{\linewidth}{!}{
\begin{tabular}{lllllllllllll} 
\hline
\multirow{3}{*}{Method} & \multicolumn{6}{c}{\textbf{Seen Kitchens (S1)}} &  \multicolumn{6}{c}{\textbf{Unseen Kitchens (S2)}}  \\
\cline{2-13}
 & \multicolumn{3}{c}{\textbf{Top-1 accuracy}}     & \multicolumn{3}{c}{\textbf{Top-5 accuracy}}   & \multicolumn{3}{c}{\textbf{Top-1 accuracy}}     & \multicolumn{3}{c}{\textbf{Top-5 accuracy}}   \\ 
\cline{2-13}
                & VERB           & NOUN           & ACT.           & VERB           & NOUN           & ACT.   & VERB           & NOUN           & ACT.           & VERB           & NOUN           & ACT.          \\ 
\hline
RU-LSTM~\cite{furnari2020rolling}              & 33.04          & 22.78          & 14.39          & 79.55          & 50.95          & 33.73  & 27.01          & 15.19          & 08.16          & 69.55          & 34.38          & 21.10           \\ 
\hline
Lat. Goal~\cite{Roy_2022_WACV}            & 27.96          & 27.40          & 8.10           & 78.09          & 55.98          & 26.46      & 22.40          & 19.12          & 04.78          & 72.07          & \textbf{42.68} & 16.97           \\
\hline
SRL~\cite{qi2021self}                  & 34.89          & 22.84          & 14.24          & 79.59          & 52.03          & 34.61        & 27.42          & 15.47          & 08.88          & 71.90          & 36.80          & 22.06             \\ 
\hline
ImagineRNN~\cite{wu2021Anticipate}           & 35.44          & 22.79          & 14.66          & 79.72          & 52.09          & 34.98       & 29.33          & 15.50          & 09.25          & 70.67          & 35.78          & 22.19            \\ 
\hline
Temp. Agg.~\cite{sener2020temporal}           & 37.87          & 24.10          & 16.64          & 79.74          & 53.98          & 36.06      & 29.50          & 16.52          & 10.04          & 70.13          & 37.83          & 23.42      \\ 
\hline
MM-Trans~\cite{roy2021action}             & 28.59          & 27.18          & 10.85          & 78.64          & 57.66          & 30.83        & 26.80          & 18.40          & 06.76          & 70.40          & 44.18          & 20.04        \\ 
\hline
MM-TCN~\cite{zatsarynna2021multi}               & 37.16          & 23.75          & 15.45          & 79.48          & 51.86          & 34.37     & 30.66          & 14.92          & 08.91          & 72.00          & 36.67          & 21.68       \\ 
\hline
AVT~\cite{girdhar2021anticipative}               & 34.36          & 20.16          & 16.84          & 80.03          & 51.57          & 36.52    & 30.66          & 15.64          & 10.41          & 72.17          & 40.76          & \textbf{24.27}       \\ 
\hline
\textbf{Abstract Goal}   & \textbf{51.56} & \textbf{35.34} & \textbf{22.03} & \textbf{82.56} & \textbf{58.01} & \textbf{38.29}  & \textbf{41.41} & \textbf{22.36} & \textbf{13.28} & \textbf{73.10} & 41.62          & 24.24  \\ 
\hline
\end{tabular}
}
\caption{Comparison of anticipation accuracy with state-of-the-art on EK55 evaluation server. ACT. is action. }
\label{epic55}
\end{table*}
\begin{table}
\resizebox{\linewidth}{!}{
\begin{tabular}{lcccccc} 
\hline
\multirow{2}{*}{\textbf{Method}} & \multicolumn{3}{c}{\textbf{Top-1 accuracy}} & \multicolumn{3}{c}{\textbf{Mean Class accuracy}} \\ 
\cline{2-7}
 & VERB & NOUN & ACT. & VERB & NOUN & ACT. \\ 
\hline
I3D-Res50~\cite{carreira2017quo} & 48.0 & 42.1 & 34.8 & 31.3 & 30.0 & 23.2~ \\ 
\hline
FHOI~\cite{liu2020forecasting} & 49.0 & 45.5 & 36.6 & 32.5 & 32.7 & 25.3~ \\ 
\hline
RU-LSTM~\cite{furnari2020rolling} & 50.3 & 48.1 & 38.6 & - & - & - \\ \hline
AVT~\cite{girdhar2021anticipative} & 54.9 & 52.2 & 43.0 & 49.9 & 48.3 & 35.2 \\
\hline
\textbf{Abstract Goal}  & \textbf{64.8} & \textbf{65.3} & \textbf{49.8} & \textbf{63.4} & \textbf{55.6} & \textbf{37.4}\\
\hline
\end{tabular}
}
\caption{Comparison on anticipation performance on EGTEA Gaze+. All methods are evaluated on fixed anticipation time of 0.5s following~\cite{girdhar2021anticipative}. 
}
\label{egtea}
\end{table}
\begin{table}
\resizebox{\linewidth}{!}{
\begin{tabular}{lcccccc}
\hline
\multirow{2}{*}{\textbf{Method}} & \multicolumn{3}{c}{\textbf{Overall}} & \multicolumn{3}{c}{\textbf{Unseen Kitchens}}\\ 
\cline{2-7}
& VERB & NOUN & ACT. & VERB & NOUN & ACT.\\
\hline
{RU-LSTM~\cite{damen2021rescaling}}  & 25.25 & 26.69 &	11.19 & 19.36 & 26.87 & 09.65\\ \hline
\begin{tabular}[c]{@{}l@{}}Temp. Agg.\\~\cite{sener2020temporal}\end{tabular} &
21.76 &	30.59 &	12.55 &	17.86 &	27.04 &	10.46\\ \hline
AVT~\cite{girdhar2021anticipative} & 26.69 & 32.33 & 16.74 & 21.03 & 27.64 & 12.89 \\ \hline
TransAction~\cite{gu2021transaction} & \textbf{36.15} & 32.20 & 13.39 & 27.60 & 24.24 & 10.05 \\ \hline
MeMViT~\cite{wu2022memvit} & 32.20 & \textbf{37.00} & \textbf{17.70} & 28.60 & 27.40  & 15.20 \\ \hline
\textbf{Abstract goal} & 31.40 & 30.10 & 14.29 & \textbf{31.36} & \textbf{35.56} & \textbf{17.34}\\ \hline
\end{tabular}  
}  
\caption{Comparison on EK100 dataset on evaluation server using test set. Accuracy measured by class-mean recall@5 (\%) following the standard protocol.}
\label{epic100}
\end{table}

{Despite, these excellent results in both EK55 and EGTEA Gaze+ datasets, our overall results on EK100 are not state-of-the-art--see Table~\ref{epic100}.
The overall performance is affected by tail classes where our method performs not as well as recent transformer methods (see Supplementary Table 1) that are heavily trained on external data~\cite{girdhar2021anticipative,wu2022memvit}.
EK100 dataset is dominated by long-tailed distribution where 228 noun classes out of 300 are in the tail classes. 
Similarly, 86 verb classes out of 97 are in the tail classes.
In our model, the next-action-representation is modeled with a Gaussian distribution (Equation~\ref{prior2}), and therefore, it is not able to cater to exceptionally long tail class distributions as in EK100.  
This is a limitation of our method. 
We do not witness the tail-class issue in EK55 as the performance measure used is accuracy compared to mean-recall in EK100. 
Accuracy is influenced heavily by frequent classes but mean-recall treats all classes equally.
However, our model shows excellent generalization results on unseen kitchens of EK100 dataset outperforming the best Transformer model~\cite{wu2022memvit}.

\subsection{Impact of goal consistency criterion and loss}
In this section, we evaluate the impact of Goal Consistency (GC) criterion and loss ($\mathcal{L}_{GC}$) using the validation set of EK55 and EK100 datasets. 
We train separate models for verb and noun anticipation using TSN-RGB (RGB) and Object (OBJ) features, respectively. 
As Mean and Median sampling are used in prior variational prediction models~\cite{abu2019uncertainty}, here we use mean and median sampling as two baselines to show the effect of GC. 
After sampling $Q \times K$ number of next-action representations ($\mathbf{a_N}$), instead of selecting the best next-action candidate using GC (Algorithm~\ref{alg}), we obtain the mean/median vector of all sampled candidates and then make the prediction using the classifier (e.g. mean vector$=\frac{\sum \mathbf{a_N}}{Q \times K}$). 
We also experimented with a majority/median class prediction baseline.
In this case, we take all $Q \times K$ predictions from the classifier (from the next action-representation candidates) and pick the majority/median class as the final prediction.
Everything else stays the same for all these mean/majority/median baseline models, except we do not use GC loss ($\mathcal{L}_{GC}$) or the GC criterion (Equation~\ref{symkld}). 
Results are reported in Table~\ref{tbl.gc}.

\begin{table}[t]
\centering
\resizebox{\linewidth}{!}{
\begin{tabular}{llcccccccc} 
\hline
\multicolumn{2}{l}{\multirow{2}{*}{\begin{tabular}[c]{@{}l@{}}Goal candidate (Q) \& \\Action~candidate (K)\end{tabular}}} & \multicolumn{4}{c}{EK55} & \multicolumn{4}{c}{EK100} \\ 
\cline{3-10}
\multicolumn{2}{l}{} & V@1 & V@5 & N@1 & N@5 & V@1 & V@5 & N@1 & N@5 \\ 
\hline
Mean &\multirow{2}{*}{\begin{tabular}[c]{@{}l@{}}\\ Q=1,\\ K=10\end{tabular}}   & 41.79 & 72.23 & 25.79 & 49.50 & 44.51 & 76.89 & 22.72 & 50.78\\
\cline{1-1}\cline{3-10}
Median & & 41.16 & 71.32 & 24.30 & 48.31 & 45.44 & 77.91 & 22.15 & 51.23\\
\cline{1-1}\cline{3-10}
Majority class & & 41.98 & 72.89 & 25.98 & 50.01 & 42.98 & 74.56 & 24.13 & 53.45 \\
\cline{1-1}\cline{3-10}
Median class & & 41.02 & 72.11 & 22.88 & 49.87 & 44.19 & 77.00 & 22.97 & 51.98\\
\cline{1-1}\cline{3-10}
Our model & & \textbf{45.18} & \textbf{77.30} & \textbf{28.16} & \textbf{51.08} & \textbf{48.84} & \textbf{80.52} & \textbf{27.50} & \textbf{55.83} \\ \hline
Mean &\multirow{2}{*}{\begin{tabular}[c]{@{}l@{}}\\ Q=3,\\ K=10\end{tabular}} & 39.40 & 72.23 & 24.22 & 48.96 & 45.90 & 77.88 & 22.41 & 50.87 \\
\cline{1-1}\cline{3-10}
Median & & 41.32 & 71.32 & 26.60 & 51.70 & 45.63 & 77.02 & 24.33 & 52.87 \\
\cline{1-1}\cline{3-10}
Majority class & & 38.39 & 69.42 & 24.70 & 48.22 & 45.72 & 78.61 & 22.61 & 50.89\\
\cline{1-1}\cline{3-10}
Median class & & 40.43 & 71.43 & 26.52 & 52.33 & 45.84 & 78.09 & 23.78 & 52.33 \\
\cline{1-1}\cline{3-10}
Our model & & \textbf{44.68} & \textbf{77.14} & \textbf{28.29} & \textbf{53.78} & \textbf{49.02} & \textbf{80.86} & \textbf{28.52} & \textbf{54.91}\\ \hline
Without $\mathcal{L}_{GC}$ & \multirow{2}{*}{\begin{tabular}[c]{@{}l@{}}Q=1,\\ K=1\end{tabular}}&38.31 & 70.77 & 19.74 & 43.11 & 43.82 & 77.45 & 21.25 & 51.99\\
\cline{1-1}\cline{3-10}
With $\mathcal{L}_{GC}$ & & \textbf{40.88} & \textbf{71.43} & \textbf{22.09} & \textbf{46.29} & \textbf{46.80} & \textbf{78.41} & \textbf{26.80} & \textbf{53.32}\\
\hline
\end{tabular}
}
\caption{\textit{The impact of goal consistency criterion and loss.} @1 and @5 denotes Top-1 and Top-5 accuracy and V stands for verb and N stands for noun.}
\label{tbl.gc}
\end{table}
As can be seen from the results, there is a significant impact of GC. 
Especially, there is an improvement of 3.39 and 2.37 for top-1 verb and noun accuracy respectively using our GC model in the EK55 dataset for $Q=1,K=10$ over Mean sampling baseline.
Similar trend can be seen for EK100 and $Q=3,K=10$ as well.
Our model also outperforms majority and median class sampling baselines for both [$Q=1,K=10$] and [$Q=3,K=10$] configurations indicating the effectiveness of goal consistency.
Overall, our method with GC loss and criterion performs better than all other variants.
Perhaps this is because the GC criterion allows the model to regularize the candidate selection while GC loss allows the model to enforce this during the training. This clearly shows the impact of \emph{goal consistency formulation} of our model for action anticipation. 
{To understand goal consistency qualitatively, we measured KLD between abstract goals of different videos. 
We found that videos with similar goals have lower KLD ($6.41 \times 10^{-5}$) than dissimilar goals ($2 \times 10^{-4}$). 
Details in Supplementary Table 2.}

We perform a more controlled experiment to further evaluate the impact of GC loss where we set $Q=1$ and $K=1$ and train our model with and without GC loss ($\mathcal{L}_{GC}$).
It should be noted that when $Q=1$ and $K=1$, there is no impact of GC criterion. 
To obtain a statistically meaningful result, we repeat this experiment 10 times and report the mean performance. As it can be seen from the results in Table~\ref{tbl.gc} (last two rows), clearly the GC loss has a positive impact even when we just sample a single action candidate from our stochastic model. 
Adding GC loss improves the top1-verb prediction by 2.57 and top1-noun prediction by 2.35 on EK55 dataset.
On EK100 dataset, the improvement is 2.98 and 4.55 for top-1 verb and noun prediction.
This clearly proves that not only does the GC criterion help, but also the GC loss improves the performance of the model.
We see that compared to our model variant [$Q=1, K=1$ with $\mathcal{L}_{GC}$] the [$Q=1, K=10$ with $\mathcal{L}_{GC}$] model performs significantly better (last row vs row 5 of Table~\ref{tbl.gc}).
This indicates the impact of next-action-representation sampling (Equation~\ref{prior2}) even for a single sampled feature-based abstract goal ($Q=1$).
We conclude that the goal consistency loss, the goal consistency criterion, and next-action-representation distribution modeling (all novel concepts introduced in this paper) are effective for action anticipation.

\begin{table}[t]
\centering
\resizebox{\linewidth}{!}{
\begin{tabular}{lccccccccc} 
\hline
\multicolumn{1}{l}{\multirow{2}{*}{parameter}} & \multicolumn{1}{l}{\multirow{2}{*}{value}} & \multicolumn{4}{c}{EK55} & \multicolumn{4}{c}{EK100} \\ 
\cline{3-10}
& & V@1 & V@5 & N@1 & N@5 & V@1 & V@5 & N@1 & N@5\\ 
\hline
\multirow{5}{*}{\begin{tabular}[c]{@{}l@{}}\textbf{num. feature-based }\\\textbf{abstract goals (Q) }\\(K = 10)\end{tabular}} 
& 1 & 45.18 & 77.30 & 28.16 & 51.08 & 48.84 & 80.52 & 27.50 & \textbf{55.83} \\
\cline{2-10}
 & 2 & 44.44 & 76.19 & \textbf{28.47} & 52.38 & 49.25 & 80.44 & 28.41 & 55.65 \\
\cline{2-10}
 & 3 & 44.68 & 77.14 & 28.29 &\textbf{53.78} & 49.02 & \textbf{80.86} & \textbf{28.52} & 54.91 \\
\cline{2-10}
 & 4 & 45.31 & \textbf{77.91} & 26.28 & 50.33 & 48.86 & 80.46 & 28.16 & 55.11 \\ 
\cline{2-10}
 & 5 & \textbf{45.80} & 77.40 & 26.95 & 51.93 & \textbf{49.71} & 80.40 & 28.04 & 55.16 \\
\hline
\multirow{6}{*}{\begin{tabular}[c]{@{}l@{}}\textbf{num. next action}\\\textbf{candidates (K)}\\(Q=3)\end {tabular}} 
& 1 & 39.81 & 72.31 & 21.48 & 44.96 & 44.24 & 75.67 & 20.06 & 42.56\\
\cline{2-10}
& 3 & 40.49 & 74.20 & 22.60 & 46.22 & 44.37 & 76.11 & 21.07 & 44.51 \\
\cline{2-10}
& 5 & 41.32 & 74.26 & 23.17 & 48.23 & 45.61 & 78.91 & 22.91 & 45.12 \\
\cline{2-10}
& 10 & \textbf{44.68} & 77.14 & \textbf{28.29} & \textbf{53.78} & 49.02 & 80.86 & \textbf{28.52} & \textbf{54.91} \\
\cline{2-10}
 & 20 & 43.79 & \textbf{79.00} & 27.07 & 51.10 & 49.01 & 80.36 & 28.13 & 55.40 \\
\cline{2-10}
 & 30 & 44.56 & 77.81 & 27.80 & 51.00 & \textbf{49.18} & \textbf{81.20} & 27.44 & 53.42 \\
\hline
\end{tabular}
}
\caption{\textit{Ablation on the sensitivity of number of sampled feature-based-abstract-goals ($Q$) and next-action representation candidate $K$ on EK55 and EK100 validation set.}}
\label{tbl.sens}
\end{table}

\subsection{Sensitivity of model on $Q$ and $K$}
Next, we evaluate how sensitive our model is to the number of sampled feature-based abstract goals $Q$ and sampled next action-representation candidates $K$. Results are shown in Table~\ref{tbl.sens}. 
Our model is not that sensitive to the sampled number of feature-based abstract goals $Q$, especially when $K>1$.
However, having $Q=3$ is better than $Q=1$ for the majority of the performance measures except for top5-noun accuracy.
As the model is not that sensitive to $Q$, we select $Q=3$ as the default.
The performance of the model increases with increasing $K$.
We select $K=10$ as our default based on results in Table~\ref{tbl.sens}.
Our model uses both GC loss and criterion during training.
GC criterion helps the model to pick the most suitable next action candidate while GC loss encourages the model to generate good candidates.  
Therefore, our model is regularized to predict the most likely candidates accurately and may not need to rely too much on the sampling process. 
We also notice that the model is not sensitive to different starting seeds in the sampling process--see Supplementary Table 4.

\begin{table}
\centering
\resizebox{0.8\linewidth}{!}{%
\begin{tabular}{lllll}
\hline
\textbf{Losses} & V@1 & V@5 & N@1 & N@5 \\
\hline
$\mathcal{L}_{NA}$ & 21.36 & 69.69 & 27.76 & 51.89 \\ 
\hline
$\mathcal{L}_{NA} + \mathcal{L}_{OG}$ & 44.42 & 77.79 & 28.41 & 51.31 \\ 
\hline
$\mathcal{L}_{NA} + \mathcal{L}_{NG}$ & 46.01 & 77.94 & 29.05 & 52.32 \\ 
\hline
$\mathcal{L}_{NA} + \mathcal{L}_{GC}$ & 43.83 & 77.43 & 28.06 & 51.87 \\ 
\hline
$\mathcal{L}_{NA} + \mathcal{L}_{OG} + \mathcal{L}_{NG}$ & 44.47 & 77.12 & 28.51 &  51.34 \\ 
\hline
$\mathcal{L}_{NA} + \mathcal{L}_{OG} + \mathcal{L}_{GC}$ & 45.47 & 77.42 & 28.61 &  52.34 \\ 
\hline
$\mathcal{L}_{NA} + \mathcal{L}_{OG} + \mathcal{L}_{NG} + \mathcal{L}_{GC}$ & \textbf{46.37} & \textbf{77.97} & \textbf{29.86} & \textbf{52.74}\\ 
\hline
\end{tabular} 
} 
\caption{\textit{Loss ablation on EK55 validation set.} $i.e. \mathcal{L}_{NA}$-Next action cross-entropy loss, $\mathcal{L}_{OG}$-Feature-based abstract goal loss, $\mathcal{L}_{NG}$-Action-based abstract goal loss, $\mathcal{L}_{GC}$-Goal consistency loss.}
\label{lossablation}
\end{table}

\begin{table}
\centering
\resizebox{0.8\linewidth}{!}{%
\begin{tabular}{lllll} 
\hline
\textbf{Model} & V@1 & V@5 & N@1 & N@5 \\
\hline
VRNN (Mean) & 27.76 & 61.23 & 22.34 & 46.78 \\ \hline
VRNN (Median) & 38.13 & 68.94 & 23.85 & 47.56 \\ \hline
\textbf{Abstract Goal}  & \textbf{44.68} & \textbf{77.14} & \textbf{28.29} & \textbf{53.78} \\ 
\hline
\end{tabular}
}
\caption{\textit{Ablation on architecture}. We compare our model with standard VRNN classification on EK55 dataset.}
\label{tbl.arch}
\end{table}

\subsection{Evaluating the impact of all loss functions}

We also study the impact of each loss function described in Section~\ref{training} and report the results in Table~\ref{lossablation}. If we use only the supervised cross-entropy loss (i.e., $\mathcal{L}_{NA}$), then the performance is the worst, especially for verbs. Both $\mathcal{L}_{OG}$ and $\mathcal{L}_{NG}$ help in regularizing the abstract goal representations ($\mathbf{z_t}$ and $\mathbf{a_N}$), and therefore results improve significantly. Especially, the $\mathcal{L}_{NA} + \mathcal{L}_{NG}$ is the best loss combination for a pair of losses. When we combine all four losses, we get the best results. While $\mathcal{L}_{NA} + \mathcal{L}_{NG}$ regularizes the learning of abstract goal representations, $\mathcal{L}_{GC}$ which minimizes the divergence between feature-based and action-based goal distributions improves the choice of next verb or noun among the plausible candidates. We conclude that all four losses are important for our model. 

\subsection{Ablation on model architecture}
We also compare our model with standard Variational RNN classification. 
We obtain the latent variable $\mathbf{z_T}$ and the observed action representation $\mathbf{a_O}$ from Equation~\ref{eq.ao}, we classify $\mathbf{a_O}$ using a classifier to obtain the future action. 
In this case, we train the VRNN model with cross-entropy loss and KL-divergence (Equation~\ref{kld_timestep}).
We call this baseline as VRNN, where we sample 30 candidates for $\mathbf{a_O}$ to match our baseline architecture with 30 next action candidates  ($Q=3, K=10$). As shown in Table~\ref{tbl.arch}, our method performs much better than the VRNN method (either considering mean or median prediction).

\subsection{Ablation on anticipation horizon }

In Table~\ref{ek55ant}, we see that our model produces consistent anticipation performance for various anticipation time horizons on EK55 dataset.
One model is trained for each anticipation duration.  
On all anticipation times, abstract goal outperforms existing approaches substantially (5-12\%).
Even when the anticipation time is increased to 2 seconds, abstract goal performance does not drop significantly compared to RU-LSTM~\cite{would} or Temp. Agg.~\cite{sener2020temporal}.
We conclude that our model can produce better predictions over longer anticipation times, due to stochastic modeling of action anticipation with goal consistency. Additional ablation studies in Supplementary Tables 3-9.
\begin{table}[t]
\centering
\resizebox{\linewidth}{!}{
\begin{tabular}{lcccc} 
\hline
\multirow{2}{*}{\begin{tabular}[c]{@{}l@{}}\textbf{Top-5 Action Accuracy}\end{tabular}} & \multicolumn{4}{c}{Anticipation Time} \\ 
\cline{2-5}
 & \multicolumn{1}{c}{2} & \multicolumn{1}{c}{1.5} & \multicolumn{1}{c}{1} & \multicolumn{1}{c}{0.5} \\ 
\hline
EL~\cite{jain2016recurrent} & 24.68 & 26.41 & 28.56 & 31.50 \\
\hline
\begin{tabular}[c]{@{}l@{}}RU-LSTM
\cite{would} \end{tabular} & 29.44 & 32.24 & 35.32 & 37.37 \\ \hline
\begin{tabular}[c]{@{}l@{}}Temp. Agg.
\cite{sener2020temporal}
\end{tabular}& 30.90 & 33.70 & 36.40 & 39.50 \\  \hline
\textbf{Abstract Goal} & \textbf{42.90} & \textbf{43.61} & \textbf{43.91} & \textbf{44.26} \\
\hline
\end{tabular}
}
\caption{Comparing Action Anticipation performance for different anticipation times on EK55 validation set. Our model consistently outperforms other approaches by a large margin for all anticipation times.}
\label{ek55ant}
\end{table}
%


\section{Limitations, Discussion, and Conclusions}
We present a novel approach for action anticipation where abstract goals are learned with a stochastic recurrent model. 
We significantly outperform existing approaches on EPIC-KITCHENS55 and EGTEA Gaze+ datasets. 
Our model generalizes to unseen kitchens in both EPIC-KITCHENS55 and EPIC-KITCHENS100 datasets, even outperforming Transformer-based models by a large margin. 
We also show the importance of goal consistency criterion, goal consistency loss, next-action representation modeling, and architecture. 
One limitation of the current work is the inability to directly interpret the abstract goal representation learned by our model. 
Second, our method is not able to tackle long-tail-class distribution issues. 
In the future, we aim to address these limitations of our model. 

\section*{Acknowledgment}
This research/project is supported in part by the National Research Foundation Singapore and DSO National Laboratories under the AI Singapore Programme (AISG Award No: AISG2-RP-2020-016) and the National Research Foundation, Singapore under its AI Singapore Program (AISG Award Number: AISG-RP-2019-010).

\bibliographystyle{plain}
\bibliography{abs_goal_arxiv}

\end{document}